\documentclass[10pt,twocolumn,letterpaper]{article}

\usepackage[pagenumbers]{cvpr} 

\usepackage[dvipsnames]{xcolor}

\definecolor{cvprblue}{rgb}{0.21,0.49,0.74}
\usepackage[pagebackref,breaklinks,colorlinks,citecolor=cvprblue]{hyperref}

\def\paperID{8} 
\def\confName{CVPR}
\def\confYear{2024}

\title{AI-EDI-SPACE: A Co-designed Dataset for Evaluating the Quality of Public Spaces}

\author{Shreeyash Gowaikar$^{1}$\thanks{Work done while interning at Chaire Unesco en Paysage Urbain.}\quad
Hugo Berard$^{2}$\thanks{Correspondance to hugo.berard@umontreal.ca}\quad
Rashid Mushkani$^{2}$\quad
Emmanuel Beaudry Marchand$^{2}$ \vspace{0.3em} \\
Toumadher Ammar$^{2}$\quad
Shin Koseki$^{2}$ \vspace{0.6em} \\
{\normalsize $^1$Birla Institute of Technology and Science} \\
{\normalsize$^2$Chaire UNESCO en Paysage Urbain, Université de Montréal}
}

\begin{document}
\maketitle
\begin{abstract}
Advancements in AI heavily rely on large-scale datasets meticulously curated and annotated for training. However, concerns persist regarding the transparency and context of data collection methodologies, especially when sourced through crowdsourcing platforms. Crowdsourcing often employs low-wage workers with poor working conditions and lacks consideration for the representativeness of annotators, leading to algorithms that fail to represent diverse views and perpetuate biases against certain groups. To address these limitations, we propose a methodology involving a co-design model that actively engages stakeholders at key stages, integrating principles of Equity, Diversity, and Inclusion (EDI) to ensure diverse viewpoints. We apply this methodology to develop a dataset and AI model for evaluating public space quality using street view images, demonstrating its effectiveness in capturing diverse perspectives and fostering higher-quality data.
\end{abstract}    

\section{Introduction}
\label{sec:intro}
Current advancements in AI heavily rely on the availability of large-scale datasets meticulously curated and annotated for training purposes. The significance of such datasets has been underscored by the success of models like ChatGPT, which leverages Reinforcement Learning with Human Feedback (RLHF) to fine-tune models based on human input \citep{achiam2023gpt}. However, concerns persist regarding the transparency and context of data collection methodologies, particularly in instances where annotations are sourced through crowdsourcing platforms. For instance, reports indicate that annotations for training ChatGPT were gathered from workers in Kenya under conditions of low pay and poor labor standards \citep{perrigo2023openai}. This reliance on crowdsourcing, often driven by cost-effectiveness, perpetuates the invisibility and exploitation of workers, particularly those from the global south \citep{gray2019ghost}.

Moreover, the failure to acknowledge the socio-cultural context within which data is produced can introduce biases into datasets. For example, algorithms trained on datasets devoid of the historical context of segregation may inadvertently perpetuate biases against certain minority groups \citep{jeff_larson_how_nodate}. Furthermore, the identities of workers involved in annotations are frequently overlooked, leading to a lack of diversity in viewpoints captured within datasets. This bias is compounded by the common practice of aggregating annotations through majority voting \citep{davani2022dealing}.

To address these limitations, we propose a methodology grounded in a specific socio-cultural context for dataset collection and AI model development. Our approach centers on a co-design model that actively involves stakeholders at key stages of the AI model development, including dataset creation. Additionally, we integrate principles of Equity, Diversity, and Inclusion (EDI) to ensure diverse viewpoints are represented within the dataset. We argue that this approach not only mitigates biases within datasets but also fosters the creation of higher-quality data reflecting diverse perspectives.

We apply this methodology to the development of a dataset and AI model capable of evaluating the quality of public spaces using street view images. Assessing public space quality is inherently subjective, as demonstrated by research showing variations across cultural groups \citep{lussault2009lutte,enos2017space}. Leveraging our proposed methodology grounded in co-design and EDI principles, we curated a dataset of streetview images annotated by a diverse group of citizens. Using this dataset, we trained a baseline AI model to score public space images along various dimensions. Finally, we propose several fairness metrics to assess the model's ability to capture diverse viewpoints within the population.

\begin{figure*}[h]
\centering
\includegraphics[width=\textwidth]{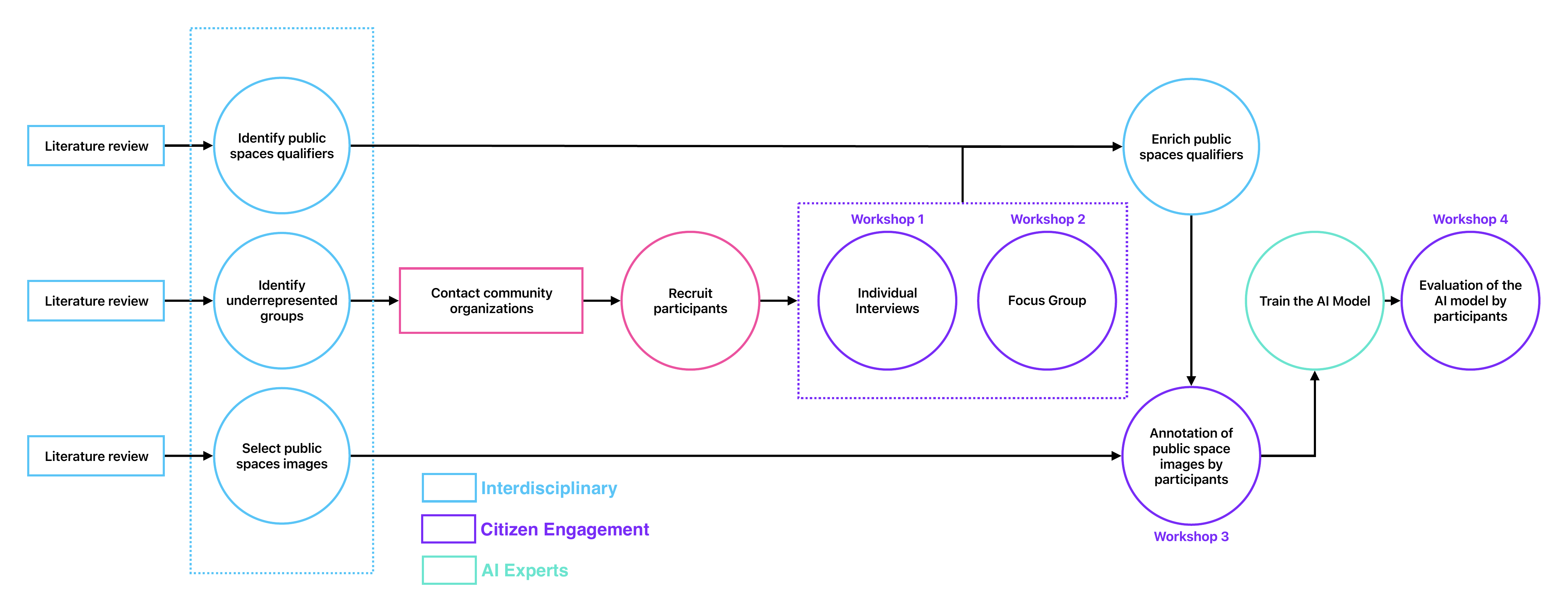}
\caption{Methodology used to create the dataset and the AI model to evaluate the quality of public spaces.}
\label{figure:methodology}
\end{figure*}

\section{Methodology}
The methodology we propose is grounded in co-design and builds on EDI principles.

\paragraph{Co-design} also known as participatory design or co-creation, embodies an approach where stakeholders are actively engaged throughout the design process to ensure that the resulting products meet their needs and preferences \cite{asaro2000transforming}. Unlike conventional design methods, which may only involve users once a product is completed, co-design aims to integrate stakeholders at every phase of development. In the context of AI, we argue that a methodology based on co-design can mitigate bias from algorithms and harm that can result from such bias. For example, a methodology based on co-design actively rejects the practice of crowdsourcing and instead seeks to involve stakeholders directly in the annotation process. The benefits of involving stakeholders are manifold:

\begin{itemize}
\item \textbf{Iterative Process}: The design process becomes iterative, allowing for feedback and refinement based on stakeholder input.
\item \textbf{Expertise}: Stakeholders can contribute their own experiences and expertise, ensuring that the data is of high quality and representative of diverse perspectives.
\item \textbf{Inclusivity}: Co-design fosters inclusivity by integrating diverse viewpoints and voices into the design process, which can lead to more fair algorithms that better serve all members of society.
\item \textbf{Shared Ownership}: Stakeholders possess a sense of ownership over the final AI, having actively participated in its creation from inception. This sense of ownership can lead to greater trust and acceptance of the technology.
\end{itemize}

\paragraph{Equity, Diversity, and Inclusion (EDI)} principles are a foundational framework of values and practices aimed at fostering fairness, representation, and belonging within organizations, communities, and societies \cite{de2022picture}. Each component of EDI is integral to creating environments where all individuals have equitable opportunities to thrive, regardless of their background or identity. In the context of AI and dataset creation, we assert that EDI principles are paramount and should be carefully considered, particularly when selecting annotators. It is imperative to ensure that annotators represent a diverse range of experiences and viewpoints. By prioritizing diversity among annotators, we can ensure that all perspectives are comprehensively incorporated into the dataset. This inclusive approach not only enhances the richness and depth of the data but also promotes fairness and equity in the resulting algorithms and models.

\paragraph{}
Based on these principles, we have developed a methodology for creating a dataset and an AI model to evaluate the quality of public spaces using street-view images. An overview of the methodology is presented in Figure~\ref{figure:methodology}. The main phases of the methodology consist of participant recruitment, the organization of workshops to understand participants' concerns regarding public spaces and establish criteria for evaluating the quality of public spaces, the annotation of the images, and the AI model evaluation.

\paragraph{Participant Recruitment:} We focused on recruiting participants from underrepresented groups. A total of 28 participants were recruited to take part in the workshops and image annotations. Recruitment efforts targeted various community organizations representing diverse underrepresented groups. Among the participants, 20 identified as women, 5 as belonging to an ethnic minority, 2 as handicapped, 10 as members of the LGBTQ2+ community, and 2 as belonging to a religious minority see Appendix~\ref{app:participants}.

\paragraph{Identification of Evaluation Criteria:} To capture the diverse uses of public spaces, we defined 35 criteria for evaluating their quality. These criteria were identified through a two-phase process: initially through a literature review and subsequently refined through feedback and discussions obtained during individual interviews and focus groups. The full list of criteria is presented in Appendix~\ref{app:criteria}.

\section{Dataset}
To evaluate the quality of public spaces, we compiled a dataset of pairwise comparisons of street view images.

\paragraph{Images} The dataset comprises 7,833 street view images gathered from the Greater Montreal region. Sampling representative images from such a vast area poses challenges, as certain regions are more densely populated and diverse than others. To address this, we implemented a two-stage sampling strategy. Initially, we sampled a 50m by 50m grid covering the entire region and identified locations with street-view images nearby. Subsequently, we excluded locations without images within a vicinity of less than 1m. In the second stage, we randomly selected a subset of images from the remaining locations. This strategy aimed to boost the number of images sampled from regions with higher street view density, typically correlating with higher population density and more diverse public spaces.

\begin{figure}[h]
    \centering \includegraphics{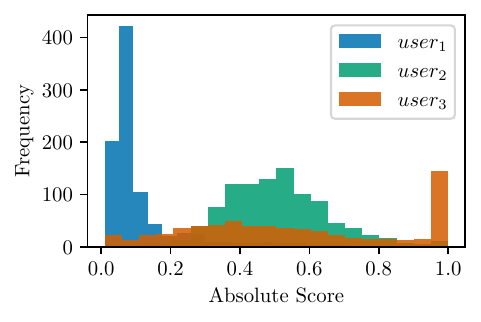}
    \caption{Voting Patterns Observed in the Streetview Dataset. It is a histogram of the absolute scores of 3 out of the 22  participants. }
    \label{voting_patterns}
\end{figure}

\paragraph{Annotations} The dataset consists of 19,990 pairwise comparisons between two images. Participants were tasked with selecting the preferred image based on a given criterion using a cursor ranging from -1 to 1. Users could indicate their preference strength by adjusting the position of the cursor, see Appendix~\ref{app:software}. Negative values denoted a preference for the left image, positive values for the right, and values close to 0 indicated no preference. This scoring method, differing from simple binary choices, mitigates Arrow's impossibility theorem \cite{arrow_difficulty_1950}, as discussed by \cite{allouah_robust_2022}. Moreover, employing continuous scores allows for quantization into finite bins during training, offering added flexibility. However, this introduces complexities in voting patterns, as illustrated in Figure~\ref{voting_patterns}.

\section{Experiments}

\subsection{Methodology}
\label{methods}

The task is defined as a pairwise Learning-to-Rank Task. 
The model indirectly learns the utility function of the user by predicting which image, amongst a pair, is given preference by the user. 
The model outputs scores for each image, and we indirectly calculate the utility by taking the difference between the scores of two images from a comparison. 

The workflow for the model is inspired by the workflow from the place-pulse 2.0 dataset \cite{dubey_deep_2016}.
The model has a feature extractor and a classifier head.
The feature extractor is a pre-trained feature extractor model.
The feature extractors we used include VGG11 \cite{simonyan_very_2014}, EfficientNet \cite{tan_efficientnet_2020}, Squeezenet \cite{iandola_squeezenet_2016}, and DinoV2 \cite{oquab_dinov2_2023}.
The features are then passed through a classifier head, which, in our case, was a single or a double-layered perceptron.
The output of the classifier is the score for each image.
We use Binary Cross Entropy, Ranking loss, and Mean Squared Error as the penalty while training the classifier.

\begin{figure*}[h]
    \centering
     \begin{subfigure}[c]{0.30\linewidth}
        \includegraphics[height =0.15\textheight]{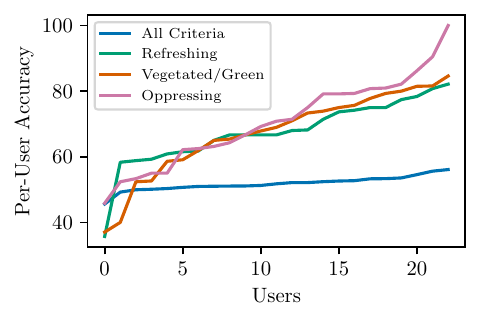}
        \caption{Sorted Per User Accuracy over i) all Criteria, ii) Top 3 Criteria - Opperessing, Vegetated/Green, Refreshing (in the descending order of accuracy).}
        \label{reg_res}
    \end{subfigure}
    \hfill
    \begin{subfigure}[c]{0.30\linewidth}
        \includegraphics[height =0.15\textheight]{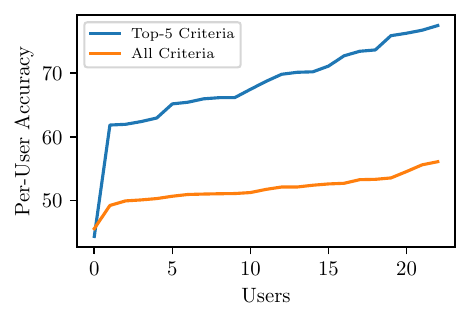}
        \caption{Sorted Per User Accuracy over i) All Criteria, ii) only 5 Criteria with highest overall accuracies (viz. Intimate, Regenerative, Refreshing, Vegetated/Green, Oppressing). }
        \label{top_5_criteria_res}
    \end{subfigure}
    \hfill
    \begin{subfigure}[c]{0.30\linewidth}
        \includegraphics[height =0.15\textheight]{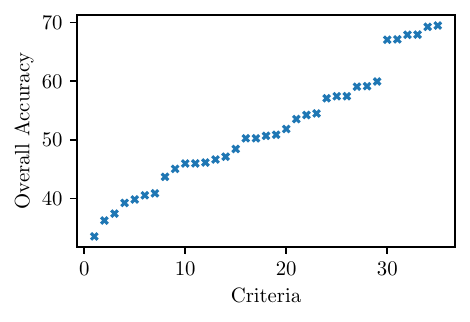}
        \caption{Scatter Plot of accuracies for every criterion. The highest accuracy is for the criterion 'Opperessing', and the lowest accuracy is for the criterion 'Inviting/Welcoming'.}
        \label{criteria_res}
    \end{subfigure}
    \caption{Experiment results}
    \label{fig_results}
    
\end{figure*}

\subsection{Equity Metrics}

To define equity for a learning-to-rank problem with respect to the participants' domain, we can view the problem as a generative problem, where the participants contribute to generating the final ranking.
The inspiration has been drawn from current equal-opportunity fairness metrics and also the Gini coefficient \cite{do_optimizing_2022}, which is widely used in economics to measure the inequity in wealth distribution in society.

\begin{itemize}
    \item Maximal Per-User Accuracy:
    \begin{equation}
        Acc_{\max} =
        \max_{i,j}(Acc_i - Acc_j)
    \end{equation}
    Here, $Acc_i$ and ${Acc}_{j}$ are the accuracies with respect to the i’th and j’th users. 
    This metric aims to calculate the maximum difference in the per-user accuracy of two users. 
    The higher the maximal per-user accuracy, the more the model has been biased towards one user than the other and hence, the user with the minimum per-user accuracy has not been given equal opportunity to affect the final ranking. 

    \item Standard Deviation of Per-User Accuracy:
    \begin{equation}
        Acc_{std}= \sqrt{\frac{\sum_{i}{(Acc_i-\bar{Acc})}}{N}}
    \end{equation}
    Here, $Acc_i$ is the accuracy with respect to the i’th user, $\bar{Acc} = \frac{1}{N}\sum_{i=1}^{N} Acc_i$ is the average per-user Accuracy, and N is the number of participants. 
    The metric aims to calculate the deviation with respect to all the per-user accuracies. 
    A large standard deviation will imply an unequal distribution of per-user accuracies, which will imply that users are not being provided with equal opportunity to affect the rankings. 

    \item Gini Coefficient over Per-User Accuracy:
    The ratio of the area underneath the Line of Equality (all the users having the same per-user accuracy) and the area underneath the Lorenz Curve (given by the cumulative per-user accuracy).

    According to the classical definition of the Gini Coefficient with respect to the relative mean difference:
    \begin{equation}
        Gini = \frac{\sum_{i, j}{(|Acc_{i} - Acc_{j}|)}}{2 N^2 \bar{Acc}}
    \end{equation}
    Here, $Acc_i$ is the accuracy with respect to the i’th user, $\bar{Acc} = \frac{1}{N}\sum_{i=1}^{N} Acc_i$ is the average per-user Accuracy, and N is the number of participants. 
    The metric aims to calculate the inequality in the model's prediction. Hence, defining Gini inequality over the per-user accuracy helps us define the inequity in the model.

\end{itemize}

\subsection{Results}
We have considered the Learning-To-Rank task on the Streetview Image Dataset as a Regression Problem, where the model outputs scores for both the images in the comparison and is trained to match the difference in the predicted score to the score provided by the participant for the particular comparison.
The value for the metrics for the regression problem type can be seen in Figure \cref{reg_res}.
Additionally, the inequity seems to be higher in the case of criteria where the model seems to perform the best overall, as seen in \cref{top_5_criteria_res} and \cref{criteria_res}. 
Here, the inequity seems to be brought about majorly by the voting patterns as the mean squared error (MSE) used to train the model penalises the model for every minute discrepancy with the users’ comparisons. 
The users with voting patterns with the majority of votes clustered around 0 (a conservative voting approach) tend to have more accuracy, as is expected for the best-fit model enforced by the MSE loss. 
However, given a different model, this learned voting pattern can differ. 
Also, the number of comparisons still affects the inequity, but it is difficult to disentangle its effects and differentiate the minor trend from the major trend of inequity by voting patterns.

\section{Conclusion}
\label{concl}

In conclusion, our study introduces a novel dataset aimed at assessing the quality of public spaces using street-view images. This dataset is the product of a methodology integrating co-design and Equity, Diversity, and Inclusion (EDI) principles, ensuring representation of diverse perspectives. We trained a baseline model on this dataset and assessed its fairness in capturing a broad spectrum of viewpoints. However, our analysis revealed significant challenges. The model's performance varied considerably across different evaluation criteria, with some criteria showing performance close to random, underscoring the complexity of the task. Moreover, we observed substantial variability in model performance across different users, indicating an inability to accurately capture preferences from diverse user groups. While these initial findings are promising, they also underscore the need for further research to develop models capable of effectively capturing and representing a diversity of viewpoints. We have tried some plausible solutions to attenuate the aforementioned problems and report their findings in \cref{apndx:sols}. Addressing these challenges is crucial for advancing the development of responsible AI models and datasets in the realm of public space evaluation.

{
    \small
    \bibliographystyle{ieeenat_fullname}
    \bibliography{main}
}

\clearpage
\appendix
\onecolumn
\section{Participants}
\label{app:participants}
\begin{figure}[h]
    \centering
    \includegraphics[width=\textwidth]{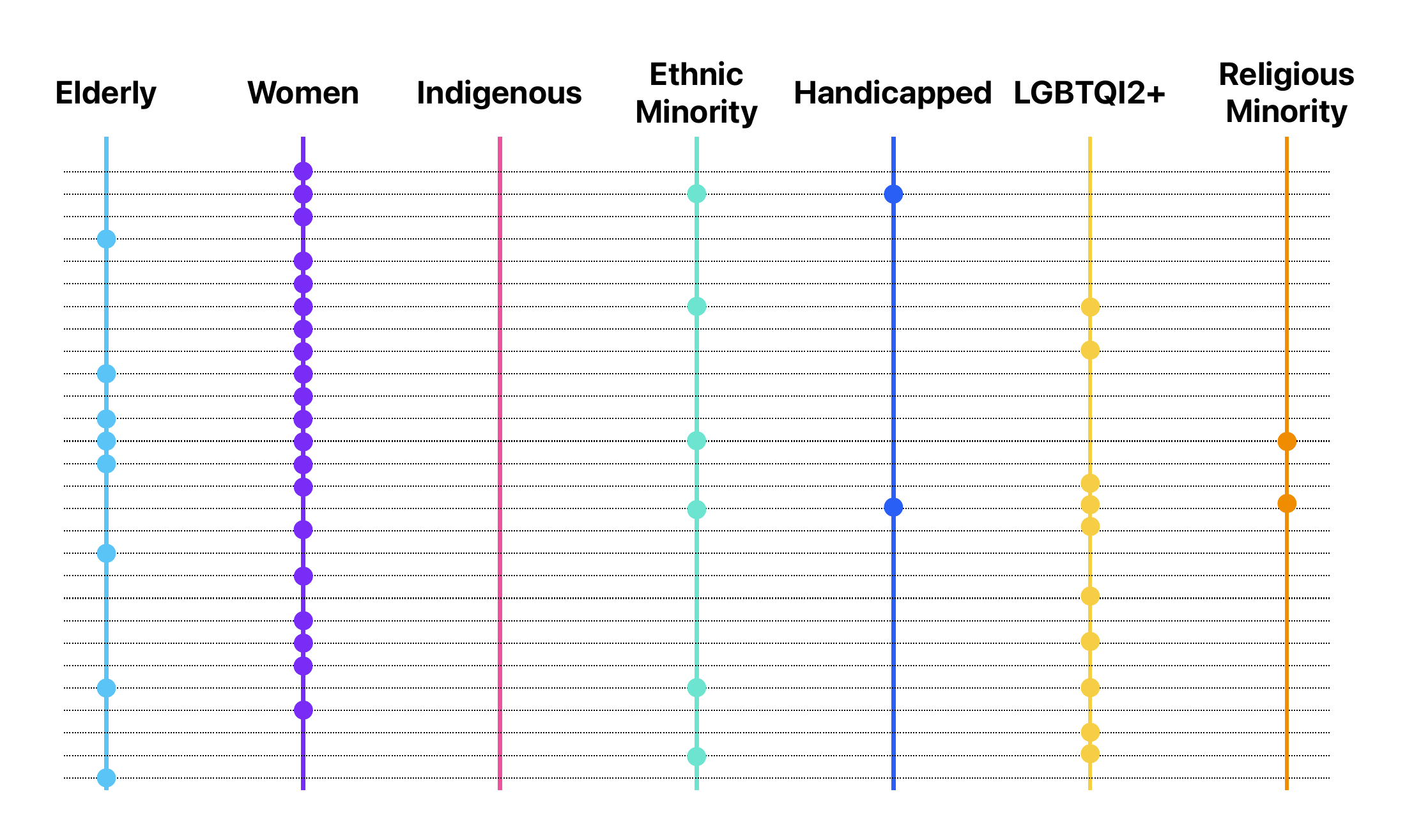}
    \caption{The identity markers of the participants.}
    \label{fig:participants}
\end{figure}

\section {Evaluation Criteria}
\label{app:criteria}
\begin{figure}[h]
    \centering
    \includegraphics[width=\textwidth]{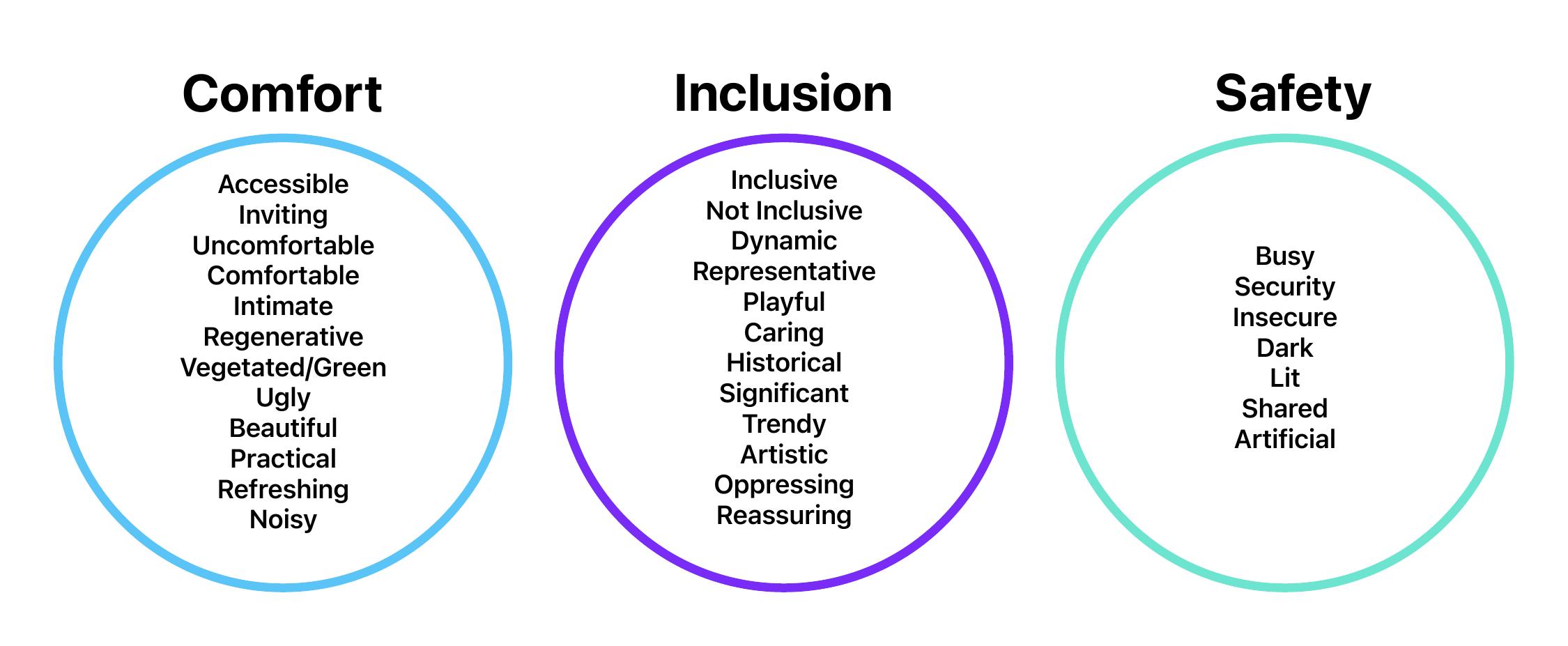}
    \caption{Criteria used for evaluating the streetview images of public spaces. }
    \label{fig:criteria}
\end{figure}

\section {Annotation User Interface}
\label{app:software}
\begin{figure}[h]
    \centering
    \includegraphics[width=\textwidth]{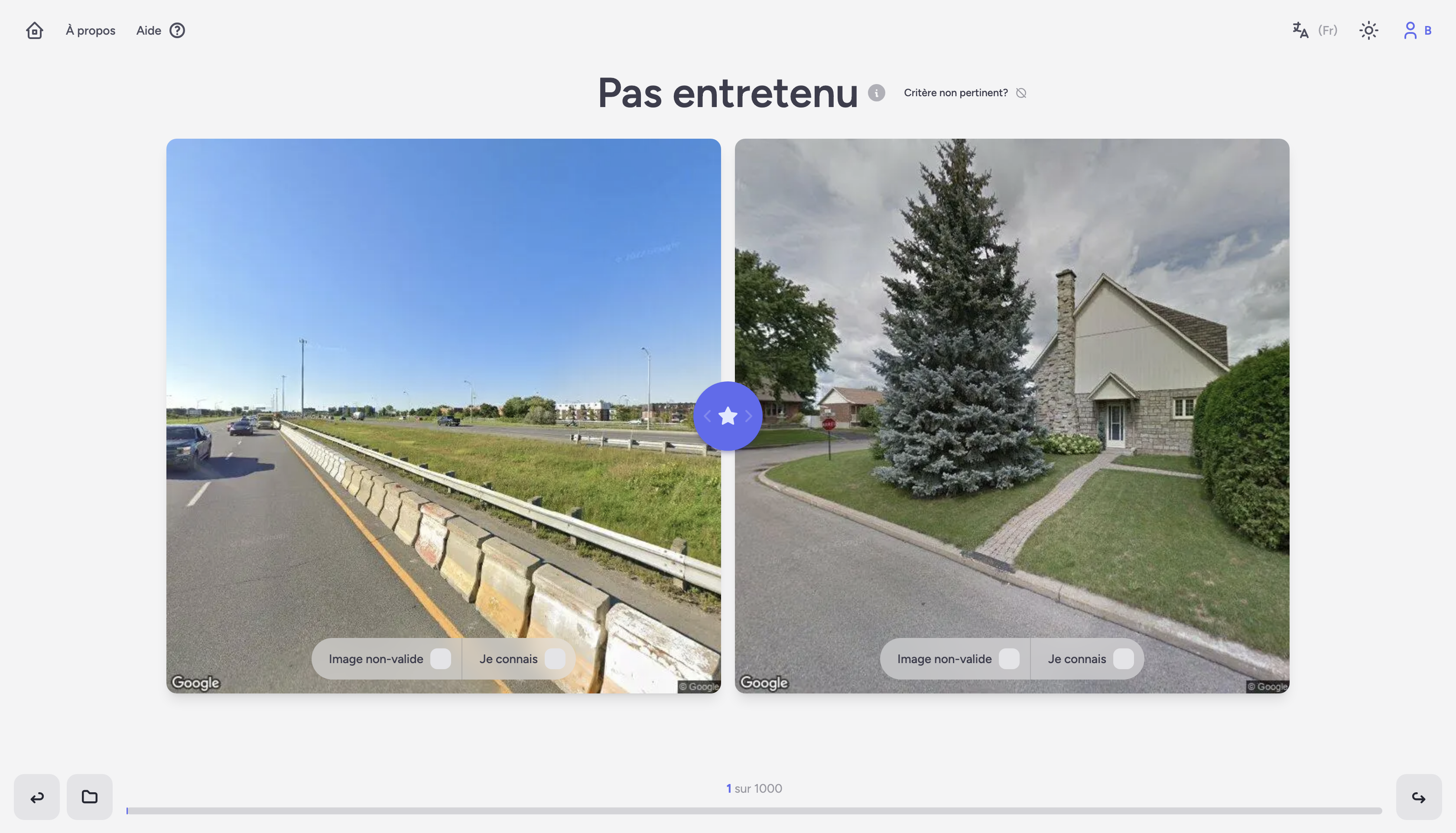}
    \caption{Screenshot of the user interface used for gathering the annotations.}
    \label{fig:software}
\end{figure}

\section{Additional Results}
\label{apndx:sols}
\begin{table}[htbp]
\begin{center}
\vskip 0.15in
\begin{small}
\begin{sc}
\begin{tabular}{lccc}
\toprule
Name of Experiment & Accuracy & Min-Max Gap  & Standard Deviation  \\
\midrule
Baseline & 47.97\%   & 7.43\%    & 2.01\%  \\
Normalisation scaling & 48.82\%   & 14.65\%    & 2.62\%  \\
User Embeddings & 49.67\%   & 5.47\%    & 1.43\% \\ 
\bottomrule
\end{tabular}
\caption{Results of our metrics over different plausible solutions}
\label{sol_results}
\end{sc}
\end{small}
\end{center}
\vskip -0.1in
\end{table}

\section{Experimentation Details}
We have implemented all experiments in this work using PyTorch Lightning. 
The model with the best performance uses an EfficientNet feature extractor, pre-trained on ImageNet, and with weights available from the TorchVision package.  
It also has a two-layered classifier with 256 hidden dimensions.
For training, we have randomly split the dataset in an 80-20 split for training and validation.
To obtain the optimal model, whose results we have presented in this work, we ran a grid search over the hyperparameters to identify the optimal set of hyperparameters.
We trained the Baseline model over 150 epochs when the training batch size was 32. 
The losses used while training the optimal model are the Least Squared Error Loss and the Ranking loss, as explained in the main text. 
We used the ADAM optimizer with an initial Learning Rate of 0.01, which was reduced iteratively if the model's performance did not change over 8 steps using a Scheduler.
Finally, we used early stopping to stop the model from overfitting and have stated the results with the model which attained the best performance while training.
Additionally, to alleviate the effects of stochasticity, we have replicated the training with the optimal hyperparameters five times, with different initial seeds.
We have trained all models on 2 Nvidia V100SMX2 GPUs, from the Beluga Computer Cluster at ETS, Montréal, with access supported through the "Digital Research Alliance of Canada (the Alliance)".

\end{document}